\documentclass{article}

\usepackage{authblk}

\usepackage{amsthm}
\usepackage{amssymb}

\usepackage{mathtools}
\usepackage{gensymb}

\usepackage{algorithm}
\usepackage{algorithmic}
\usepackage{makecell}

\usepackage{enumitem}

\usepackage[T1]{fontenc}
\usepackage{lmodern}

\usepackage[english]{babel}

\usepackage[letterpaper,top=2cm,bottom=2cm,left=3cm,right=3cm,marginparwidth=1.75cm]{geometry}

\usepackage{amsmath}
\usepackage{graphicx}
\usepackage[colorlinks=true, allcolors=blue]{hyperref}

\title{Mastering the Game of Go with Self-play Experience Replay}

\author[1{\footnote{equal contribution. liujingbin01@corp.netease.com}}]{Jingbin Liu} 
\author[1]{Xuechun Wang} 
\affil[1]{NetEase AI Lab, Shanghai, China}

\date{}
\begin{document}
\maketitle

\begin{abstract}
The game of Go has long served as a benchmark for artificial intelligence, demanding sophisticated strategic reasoning and long-term planning. Previous approaches such as AlphaGo and its successors, have predominantly relied on model-based Monte-Carlo Tree Search (MCTS). In this work, we present QZero, a novel model-free reinforcement learning algorithm that forgoes search during training and learns a Nash equilibrium policy through self-play and off-policy experience replay. Built upon entropy-regularized Q-learning, QZero utilizes a single Q-value network to unify policy evaluation and improvement. 
Starting tabula rasa without human data and trained for 5 months with modest compute resources (7 GPUs), QZero achieved a performance level comparable to that of AlphaGo.
This demonstrates, for the first time, the efficiency of using model-free reinforcement learning to master the game of Go, as well as the feasibility of off-policy reinforcement learning in solving large-scale and complex environments.
\end{abstract}

\section{Introduction}
The field of Artificial Intelligence has witnessed remarkable progress driven by Deep Reinforcement Learning (RL). From the early breakthroughs in Atari games \cite{dqn}, to the landmark mastery of the game of Go by AlphaGo \cite{AlphaGo}, and the subsequent scaling to complex multi-agent environments like Dota 2 \cite{dota5} and StarCraft II \cite{alphastar}, RL has proven to be a robust framework for decision-making. More recently, the principles of RL have been pivotal in enhancing the reasoning capabilities of Large Language Models \cite{rlhf, dsr1}.

Among these benchmarks, the game of Go stands out as a unique challenge that has been solved  only with model-based RL, whereas the others primarily relied on model-free RL.
AlphaGo and its improved versions, AlphaGo Zero \cite{alpahgozero} and AlphaZero \cite{alpahzero}, demonstrated that deep RL combined with model-based MCTS could achieve superhuman performance.
Subsequently, a series of replication studies emerged. Notably, ELF OpenGo \cite{Elfopengo} clarified the training process and hyperparameters of AlphaZero, and KataGo introduced  several improvements to the AlphaZero process and architecture \cite{katago}.

Although MCTS can serve as a powerful policy improvement operator during training,  its dependence on an environment model greatly limits its generality. In this work, we propose a new algorithm, QZero, as an alternative to AlphaGo and its successors. 
On one hand, QZero is model-free and does not require an environment model, ensuring its generality; on the other hand, it is both data- and compute-efficient, enabling it to tackle the intrinsic high complexity of the game of Go.

Built upon entropy-regularized Q-learning, QZero unifies policy evaluation and improvement with a single Q-value network, and it can be trained with off-policy experience replay. It has long been believed in the RL community that value-based off-policy RL is not capable of solving large-scale complex environments, due to the severe target bias problem of Q-value backup during training. We argue that this statement is not actually true. 
In the field of model-free RL, value-based and policy-based methods can be unified under entropy regularization \cite{pgq,sqn}. Based on this observation, we explored value-based RL and achieved large-scale off-policy RL. 
When applied to the game of Go, QZero can achieve performance comparable to AlphaGo, even under limited computational resources.

We identify 3 key components essential to QZero's success:
\textbf{(i).} Ignition Mechanism.  
We propose an ignition stage to initiate Q-learning by utilizing episode returns as Q-value targets.
\textbf{(ii).} Entropy Regularization. 
We introduce entropy regularization to achieve more efficient training by smoothing the learning process to the Nash equilibrium policy.
\textbf{(iii).} Polyak Averaging. 
Unlike AlphaGo, which trains its neural networks via supervised learning, QZero's neural network training and policy improvement are intertwined, making the training process more difficult. We make use of target network with Polyak averaging to stabilize training.

The remainder of this paper is structured as follows: Section 2 details the methodology of QZero. Section 3 presents our experimental setup, training details and results. Section 4 is for discussion, and Section 5 concludes.

\section{Methods}

The approach of QZero is completely different from that of AlphaGo \cite{AlphaGo}. The core of a reinforcement learning algorithm is its policy improvement. AlphaGo achieves policy improvement through MCTS, which requires a dynamic model of the environment. QZero does not have this restriction, which makes it a more general algorithm.

\subsection{Model-free reinforcement learning}

Reinforcement learning is concerned with how an intelligent agent should take actions in a dynamic environment in order to maximize the expected cumulative reward. 

Mathematically, RL problems are often framed as a Markov Decision Process (MDP). An MDP is defined by a tuple  \((\mathcal{S,A,P,R, \gamma})\): 
\begin{itemize}
\item
  \(\mathcal {S}\) represents state space,
\item
  \(\mathcal {A}\) represents action space, 
\item
  \(\mathcal {P:S\times A \times S\to \mathbb{R}}\) stands for the transition probability distribution \(p(s^{\prime}|s,a)\) from state-action pair to next state,
\item
  \(\mathcal {R}:\mathcal{S\times A }\to \mathbb{R} \) corresponds to the
  reward function \(r(s, a)\)
\item
  \(\gamma \in (0, 1] \) determines how much the agent cares about future rewards compared to immediate rewards.
\end{itemize}
In the RL framework, the agent interacts with the environment according to a policy $a \sim \pi(\cdot|s)$, then the environment transitions to next state with $s^{\prime} \sim p(\cdot|s, a)$ and provides a reward $r$. We use the trajectory $ \tau $  to represent the path left by the agent through the state space while interacting with the environment. 
The goal of the agent is to learn an optimal policy that:
\begin{equation}
    \pi^* = \arg \underset{\pi}\max \; \mathbb{E}_{\tau \sim \pi} \left[{ \sum_{t=0}^{\infty} \gamma^t  r(s_t, a_t) } \right].
\end{equation}

Whether to use a transition model or not is what distinguishes between model-based RL and model-free RL. In model-free RL, the agent learns strictly through trial and error, where the environment transition probabilities are implicitly embedded in the experience data.
Model-free RL is generally categorized into 2 main approaches:

\textbf{Value-based methods.}
In value-based methods, the agent does not learn a policy directly. Instead, it learns the Q-value function (Q-Learning).
\begin{equation}
    Q^{\pi}(s,a) = \mathbb{E}_{\tau \sim \pi} \left[ { \sum_{t=0}^{\infty} \gamma^t   r(s_t, a_t)   \bigg| s_0 = s, a_0 = a} \right].
    \label{eq:q}
\end{equation}
The policy is usually implicitly defined as ``pick action with high Q-value.''  

Standard value-based methods include DQN \cite{dqn} and DDPG \cite{ddpg}.

\textbf{Policy-based methods.}
Here, the agent ignores the Q-value function and attempts to optimize the policy directly. 
The agent adjusts the parameters of the policy network to maximize the expected cumulative reward using gradient descent (Policy Gradient). 

The basic algorithm is called Monte-Carlo Policy Gradient (REINFORCE). In order to reduce the variance in policy-based methods, a value function is introduced as an auxiliary in actor-critic methods. Actor and critic stand for policy and value, respectively. The critic provides a more stable feedback signal to the actor than raw rewards used in pure policy-based methods. Actor-critic methods like Trust Region Policy Optimization (TRPO) \cite{trpo} and Proximal Policy Optimization (PPO) \cite{ppo} are commonly used.

\subsection{Unification under entropy regularization}

Entropy regularization is an important idea that aims at stabilizing training through better exploration in reinforcement learning \cite{sql}. 
Practical studies indicate that entropy regularization can effectively improve training efficiency and the performance of the learned policy. Entropy regularization introduces an intrinsic policy entropy reward to the original reward, and the objective of the original RL problem changes to:
\begin{equation}
    \pi^* = \arg \underset{\pi}\max \; \mathbb{E}_{\tau \sim \pi} \left[ { \sum_{t=0}^{\infty}  \gamma^t \bigg(  r(s_t, a_t) + \gamma \alpha  H\left(\pi(\cdot|s_{t+1})\right) \bigg) } \right].
    \label{eq:srl}
\end{equation}

Here, we refer to entropy regularized reinforcement learning as Soft Reinforcement Learning (SRL). It is easy to notice that SRL reduces to RL when the temperature $\alpha=0$. Every reinforcement learning algorithm has its soft counterpart. For value-based methods, DQN and DDPG's soft counterparts are Soft Q Network (SQN) \cite{sqn} and Soft Actor Critic (SAC) \cite{sac}, respectively. For policy-based methods, we have soft versions of algorithms like Soft TRPO, Soft PPO, etc \cite{sppo}. 
Furthermore, as pointed out in \cite{pgq,sqn}, soft value-based methods (Soft Q Learning) and soft policy-based methods (Soft Policy Gradient) are equivalent.

In summary, model-free reinforcement learning is unified under entropy regularization. The unification is two-fold:
\textbf{(i).}
RL is a special case of SRL. 
\textbf{(ii).}
Model-free RL algorithms are essentially equivalent in the framework of SRL.

\subsection{QZero}

Due to the unification of model-free reinforcement learning, we set our goal 
to devise a model-free RL algorithm that is as simple as possible while still being capable of handling complex environments, specifically the game of Go in this work. 

\begin{figure}[h]
\centering
\includegraphics[width=0.9\linewidth]{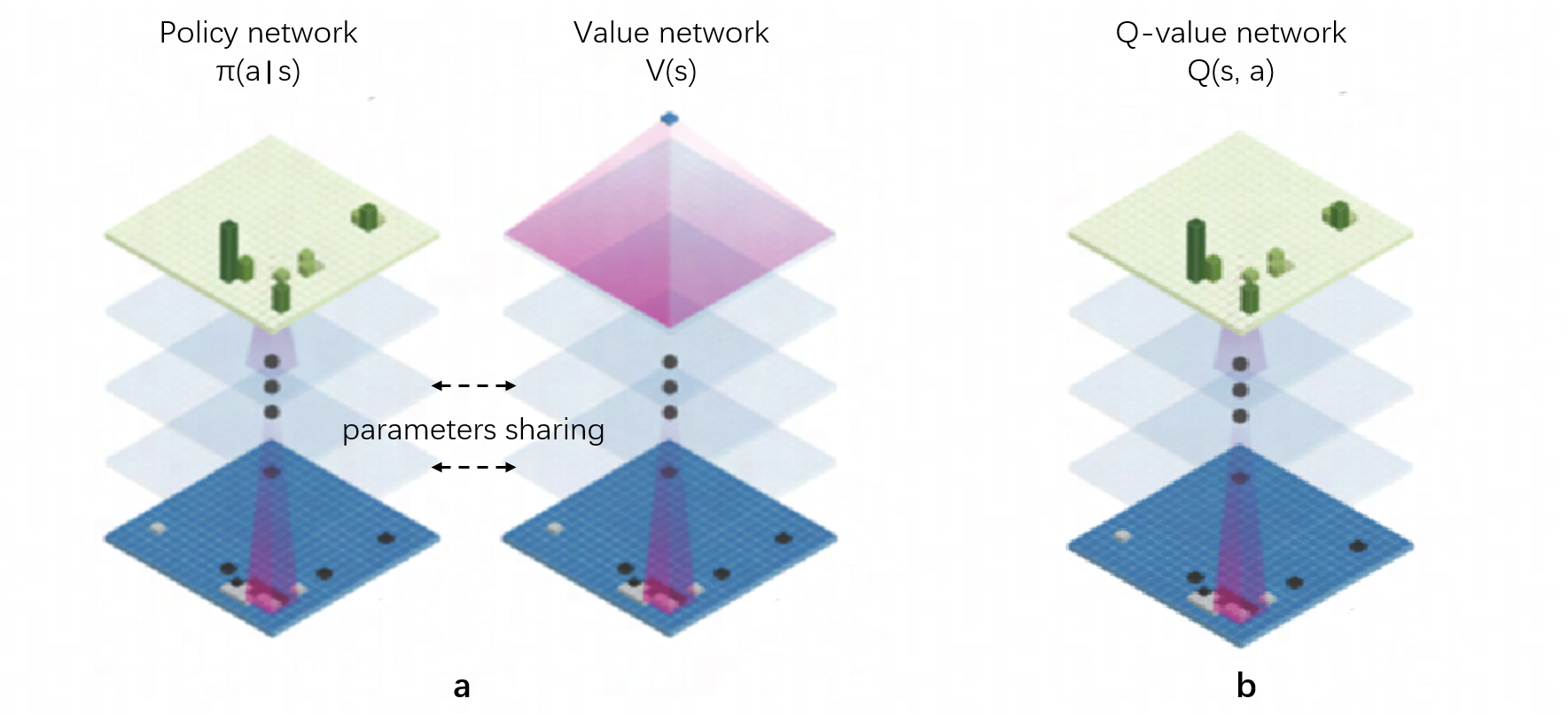}
\caption{\label{fig:nets1}Neural network architectures of AlphaGo and QZero. \textbf{a.} AlphaGo uses two networks to represent policy and value, respectively. The two networks can be configured to either share network backbone parameters or be independent. \textbf{b.} QZero uses only one network to represent Q-value. Policy and value can be derived from the single network.}
\end{figure}

QZero takes soft Q-learning as a starting point. It inherently supports data-efficient off-policy learning while requiring only one Q-value network, obviating the need for two separate networks to represent policy and value (see Figure~\ref{fig:nets1}, adapted from \cite{AlphaGo}).
Compared to a single Q-network, having both policy network and value network involves redundancy of information.

The game of Go is a standard two-player zero-sum game with perfect information.
Solving the game of Go can be approached from two perspectives: 
\textbf{(i).} seeking Nash equilibrium strategies through self-play, and 
\textbf{(ii).} bootstrapping the entire state-action space tree from terminal states back to initial states.
QZero perfectly integrates these two perspectives into its algorithm. It treats the Go problem as a unified whole; it does not distinguish between the two players or handle their strategies separately, but instead uses a single Q-value function to estimate the entire gameplay process.
We define $R_t$ as the reward of an agent at step $t$: 
\begin{equation}
    R_t = \gamma^t  ( r(s_t, a_t) - \gamma \alpha  H\left(\pi(\cdot|s_{t+1})\right) ).
    \label{eq:a1}
\end{equation}
Then, the Q-value of policy $\pi$ in the game can be expressed as:
\begin{equation}
    Q^{\pi}(s,a) = \mathbb{E}_{\tau \sim \pi} \left[ { \sum_{t=0}^{\infty} (R_{2t} -R_{2t+1} ) \bigg| s_0 = s, a_0 = a} \right].
    \label{eq:a2}
\end{equation}
Expanding $Q^{\pi}(s,a)$ with $R_t$, we have ($\gamma$ is omitted for simplicity):
\begin{align}
    Q^{\pi}(s,a) &= \mathbb{E}_{\tau \sim \pi} \Big[ {  r(s_0, a_0) - \alpha  H\left(\pi(\cdot|s_{1})\right) - r(s_1, a_1)  +  \alpha  H\left(\pi(\cdot|s_{2})\right) + \cdot\cdot\cdot \big| s_0 = s, a_0 = a} \Big]
    \label{eq:a3} \\
    &=  r(s, a) - \alpha  H\left(\pi(\cdot|s^{\prime})\right)  - \mathbb{E}_{\tau \sim \pi} \Big[ {  r(s_1, a_1)  -  \alpha  H\left(\pi(\cdot|s_{2})\right) - \cdot\cdot\cdot \big| s_1 = s^{\prime}} \Big]
    \label{eq:a5} \\
    &=  r(s, a) - \alpha  H\left(\pi(\cdot|s^{\prime})\right)  - \mathbb{E}_{\tau \sim \pi} \Big[ Q^{\pi}(s_1, a_1) \big| s_1 = s^{\prime} \Big]
    \label{eq:a6} \\
    &=  r(s, a) - \sum_{a^\prime\in \mathcal {A}} \pi  (a^\prime \mid s^\prime) \left( Q^\pi(s^\prime, a^\prime)    -\alpha  \, \text{log} \, \pi  (a^\prime \mid s^\prime)\right),
    \label{eq:a7}
\end{align}
which is the one-step backup equation of $Q^{\pi}(s,a)$. 

In the QZero algorithm, Q-value estimation is updated with the backup equation, and the policy is iterated as the softmax form over the learned Q-value. The update loop of Q-value and the policy forms the core of the algorithm. The pseudo-code of QZero is presented in Algorithm~\ref{alg:QZero}.

\begin{algorithm}[ht]
\small
\caption{\textbf{QZero}}
\label{alg:QZero}
\begin{algorithmic}[1]
    \STATE Input: initial Q-function parameters $\phi$, empty replay buffer $\mathcal{D}$
    \STATE Set target parameters equal to main parameters  $\overline{\phi} \leftarrow {\phi}$
    $$ \pi_{{\phi}} ({\cdot} \mid s) =  \text{Softmax}  \left( \frac{1}{\alpha}  Q_{{\phi}}(s, \cdot)  \right) $$
    \REPEAT
        \STATE Observe state $s$ and select action $a\sim \pi_{\overline{\phi}}(\cdot \mid s)$
        \STATE Execute $a$ in the environment
        \STATE Observe next state $s^\prime$, reward $r$, and done signal $d$ to indicate whether $s^\prime$ is terminal
        \STATE Store $(s, a, r, s^\prime, d)$ in replay buffer  $\mathcal{D}$
        \STATE Reset the environment if $s^\prime$ is terminal
        \IF{it is time to update}
            \STATE Update Q-function with episode return if it is Ignition stage
            \FOR{step in range (however many updates)}
                \STATE Randomly sample a batch of transitions, $B = \{(s, a, r, s^\prime, d)\}$ from $\mathcal{D}$
                \STATE Compute target for Q-function:
                $$y_q(r, s^\prime, d)= r - \gamma (1-d) y_v(s^\prime)$$
                 $$ y_v(s^\prime) =  \sum_{a^\prime \in \mathcal {A}} \pi_{\overline{\phi}}  (a^\prime \mid s^\prime) \left( Q_{\overline{\phi}}(s^\prime, a^\prime)    -\alpha  \, \text{log} \, \pi_{\overline{\phi}}  (a^\prime \mid s^\prime)\right) $$
                \STATE Update Q-function by one step of gradient descent using:
                $$\bigtriangledown_{\phi} \frac{1}{\mid B \mid} \sum_{ (s, a, r, s^\prime, d) \in	 B} \left( Q_{\phi}(s, a) - y_q(r, s^\prime, d) \right) ^2$$
                \STATE Update target network with Polyak averaging:
                $$\overline{\phi} \leftarrow \rho \, {\overline{\phi}} + (1-\rho) \, {\phi}$$ 
            \ENDFOR
        \ENDIF
    \UNTIL{convergence}

\end{algorithmic}
\end{algorithm}

It has been empirically challenging to stabilize off-policy Q-learning algorithms when scaling up to complex environments. A critical reason for this is the emergence of severe target bias during the Q-value learning process. Especially in the initial phase of training, the bias stemming from the completely random network parameters can result in a failure to initiate training. To address this issue, we propose an ``Ignition Mechanism'' that warms up the neural network by utilizing episode returns as targets before the standard training.

Entropy regularization, on one hand, helps smooth the state space of learning via policy entropy, while on the other hand, it introduces entropy bias. To gradually approach the optimal policy during training, we propose an $\alpha$-annealing schedule: decaying the temperature $\alpha$ over the course of training with $\alpha_i \to 0$, while maintaining a support distribution of states during self-play with $\alpha \sim \text{Uniform}(\alpha_0, \alpha_i)$.

Additionally, the update of Q-value network makes use of a target network, which has been shown to stabilize training \cite{dqn1, sac0}. Specifically, we use the Polyak averaging, which updates a neural network with an exponentially moving average of its parameters.

\section{Experiments}

We apply the QZero algorithm to the classic board game of Go. In this section, we introduce all the key components of our experiments, including the Go environment, neural network architecture, training infrastructure, training details and experiment results.

\subsection{Go environment}

In the reinforcement learning setting, 
a robust and efficient environment is fundamental for training and evaluating agents. 
In this work, we apply a Go environment that is slightly modified from GymGo \cite{GymGo}, an environment of the game Go implemented using OpenAI's Gym API. An agent interacts with the environment and generates data, which can be abstracted as a tuple $(s, a, r, s^\prime, d)$  for each timestep. Then we train the agent with the collected interaction data (see Algorithm~\ref{alg:QZero}). Details of the Go environment are described as follows:

\textbf{Observation.} 
Observation is a structured representation of the game state.
Except for action $a$, all other quantities in the tuple $(s, a, r, s^\prime, d)$ can be derived from observation. Specifically, observation is a tensor with shape (19, 19, 6), which encodes various information essential for game play \cite{GymGo}:
\begin{itemize}
\item channels 1 and 2: represent the Black and White stones, respectively.
\item channel 3: Indicator layer for whose turn it is.
\item channel 4: Invalid moves (including ko-protection) for the next action.
\item channel 5: Indicator layer for whether the previous move was a pass.
\item channel 6: Indicator layer for whether the game is over.
\end{itemize}

Based on the above game observation, we can construct features that serve as input for the neural network; the features are detailed with neural network architecture in the next subsection.
Additionally, we can also obtain action mask from the game observation.
The action space consists of 362 discrete actions: $19\times19 = 361$ board intersections plus 1 pass action. A binary action mask vector is applied to indicate illegal moves at each timestep, enforcing ko rules and preventing suicide plays. The action mask dynamically updates after each move, ensuring agents only sample from valid actions during both training and evaluation.

\textbf{Symmetry.}
The game of Go is played on a perfect square board. The observation of the environment has 8 types of symmetries, including 4 rotations by $90^{\circ}$ increments and 4 mirror reflections, which together form the dihedral symmetry group $D_4$. 
It is a natural way to enhance data diversity and improve training sample efficiency by incorporating $D_4$ symmetry augmentation. This technique generates 8 symmetric states from 1 single board position. 

However, in this work we do not use the eightfold augmented data because training data is quite abundant. Instead, in our experiment we randomly sample a symmetry operation and apply it to the original observation for every timestep. Game state and action space transform according to the symmetry operation, while scalar signals like  reward $r$ and done $d$ remain unchanged. It is crucial to exploit symmetry during training for better generalization and robustness of the learned policy.

\textbf{Game Rules and Scoring.}
The Go environment employs Tromp-Taylor scoring \cite{ttr}, a simple area-based scoring method. A player's score is defined as the sum of the number of empty points strictly surrounded by their stones and the number of their stones remaining on the board.
To balance the inherent advantage of Black moving first, a komi (compensation points) is applied to White. In this work, games are scored using Chinese rules with a komi of 7.5 points.

The game ends when both players pass consecutively, at which point the final score is calculated as (from Black's perspective):
\[
\text{score} = \text{area}_B - \text{area}_W - 7.5,
\]
where $ \text{area}_B $ and $\text{area}_W $ stand for the areas of Black and White, respectively. 
The score is then transformed into a reward function for reinforcement learning as:
\[
r = \text{sign}(\text{score}) \cdot \left( 5 + 2 \cdot \log_{10}\left(1 + |\text{score}|\right) \right),
\]
where the first item is the core reward for winning or losing; the second item is not mandatory, its introduction is aimed at encouraging active game play. Previous Go AI systems have been observed to exhibit issues with negative competition. For example, during the endgame phase, if the outcome is difficult to reverse, the AI does not compete for every point as a human player would.

As the game of Go is a standard zero-sum game, the score and reward of White are the opposites of those of Black. The game of Go is inherently a sparse reward problem, where reward signal is only available at the end of the game.

\subsection{Neural network architecture}
The neural network architecture is designed to process the observation of the environment and output the action-value estimation of the current board. Specifically, the input to the neural network is based on the feature space, while the output is based on the action space.

The input features have 2 channels: each channel is represented as a tensor with shape (19, 19), which is set to match the size the Go board. The first channel is for the current stone map and ko invalid move; the second channel is for move turn, all the values of this channel are set to 0 or 1 to indicate whose turn it is to move.
We use features that are as simple as possible in order to support a larger experience buffer size. All the features are summarized in Table~\ref{tab:feature}.

\begin{table}[h]
\centering
\begin{tabular}{l|c|l}
\hline
Feature &  channels & Description \\\hline
Stones and ko & 1 & \makecell[l]{ -1 and 1 for Black and White stones, respectively.\\0.5 for ko invalid move, 0 for others.} \\\hline
Move turn & 1 & 0 for Black, 1 for White.
\\\hline
\end{tabular}
\caption{\label{tab:feature}Features and description}
\end{table}

In this work, the neural network backbone is the same as AlphaGo Zero, except that all the normalization layers have been removed. The neural network consists of many residual blocks \cite{resnet} of convolutional layers \cite{cnn} with
rectifier non-linearities \cite{relu}. The input features are processed by a residual tower that consists of a single convolutional block followed by 19 residual blocks.

The convolutional block applies the following modules:
\begin{enumerate}[label=(\arabic*)., leftmargin=2cm]
\item A convolution of 256 filters of kernel size 3 $\times$ 3 with stride 1
\item A rectifier non-linearity
\end{enumerate}

Each residual block applies the following modules sequentially to its input:
\begin{enumerate}[label=(\arabic*)., leftmargin=2cm]
\item  A convolution of 256 filters of kernel size 3 $\times$ 3 with stride 1
\item  A rectifier non-linearity
\item  A convolution of 256 filters of kernel size 3 $\times$ 3 with stride 1
\item  A skip connection that adds the input to the block
\item  A rectifier non-linearity
\end{enumerate}

The output of the residual tower is processed by the following modules for computing Q-values:
\begin{enumerate}[label=(\arabic*)., leftmargin=2cm]
\item A convolution of 2 filters of kernel size 1 $\times$ 1 with stride 1
\item A rectifier non-linearity
\item A fully connected linear layer that outputs a vector of size $19^2 + 1 = 362$ corresponding to Q-values for all board intersections and the pass move
\end{enumerate}

\subsection{Infrastructure}
\begin{figure}[h]
\centering
\includegraphics[width=0.95\linewidth]{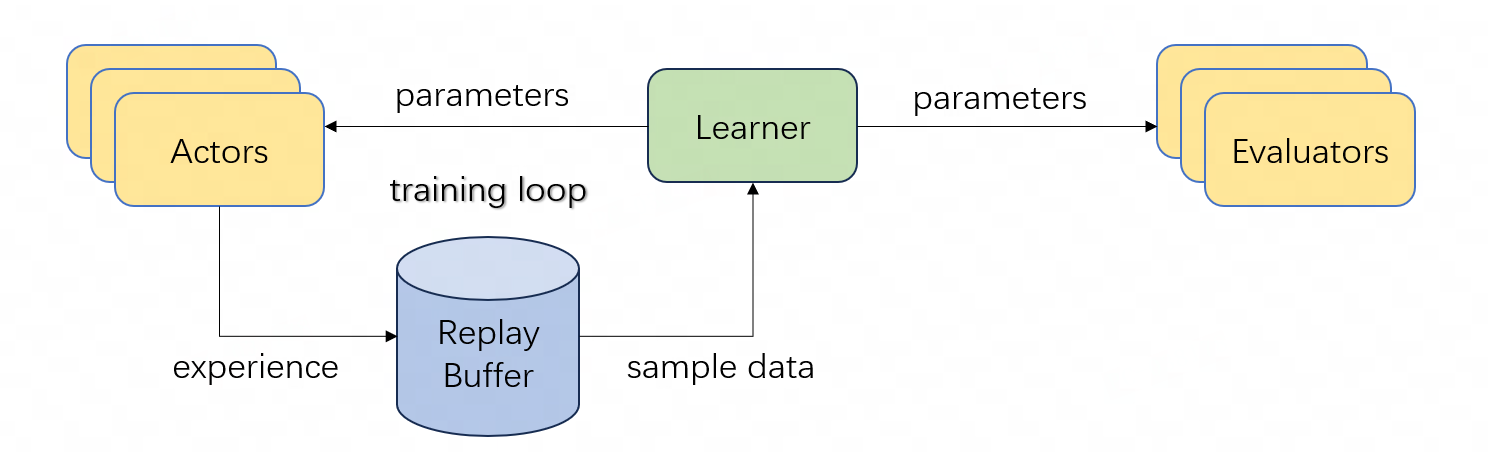}
\caption{\label{fig:arch}Training framework of QZero. Actors, Replay Buffer, and Learner constitute the training loop. Evaluators are attached to the training loop for monitoring the training process.}
\end{figure}

To better illustrate the QZero algorithm, the pseudo-code in Algorithm~\ref{alg:QZero} is written in the synchronous training paradigm. 
In our actual experiments, we use an asynchronous training version of QZero, where agent sampling and learning run concurrently to enable large-scale training.
The QZero training pipeline follows the standard off-policy reinforcement learning framework, which consists of Actors, Replay Buffer, Learner and Evaluators (see Figure~\ref{fig:arch}, adapted from \cite{deepnash}).

\begin{itemize}
\item The Actors self-play with the latest network parameters and store the $(s, a, r, s^\prime, d)$ tuples into Replay Buffer. In the Ignition phase, the final episode return $\{+5, -5\}$, which indicates winning or losing the game, is also added to the tuple.
\item The Replay Buffer, as a FIFO (first in first out) queue, stores game tuples from Actors.
\item The Learner, randomly samples a batch of game tuples from Replay Buffer, trains the neural network,
and periodically sends the latest network parameters to Actors and Evaluators.
\item The Evaluators sample network parameters from the history parameters pool, let the agent with the latest parameters play against the agent with the sampled history parameters, and obtain the game outcome for statistics.
\end{itemize}

The training pipeline is implemented on a single machine, with 1 RTX 2080Ti GPU for the self-play of Actors and 6 RTX 2080Ti GPUs for the neural network training of Learner. In comparison, AlphaZero used 5,000 TPUs to generate self-play games and 64 TPUs to train the neural networks \cite{alpahzero}.

\subsection{Training details}

We trained the Go AI agent from scratch based solely on reinforcement learning without human data. The entire project was done with 7 GPUs. To achieve stable and efficient training, we conducted some control experiments during the training process. We did not divide the GPUs to run parallel experiments. All the experiments were carried out one by one through resetting or restoring the latest checkpoint.

During the initial phase, we attempted to launch the training without the ignition mechanism. Across 3 experiments, we failed to successfully initiate training in all attempts; the neural network outputs collapsed to identical values. The training loss kept at a low level in these instances, while the neural network remained trapped in this state and failed to recover. Thereafter, we introduced the ignition phase, wherein the neural network is initially trained using episode returns instead of directly bootstrapping Q-values. We repeated the experiment 3 times with the ignition mechanism and achieved successful initialization in every instance, demonstrating that the ignition phase is both effective and essential. We believe that the ignition phase is also beneficial for other off-policy reinforcement learning algorithms.

The training process spanned 5 months. During the first 2 months, we adjusted the hyperparameters frequently, while we kept them constant for the subsequent 3 months. The temperature $\alpha$, which governs the exploration-exploitation trade-off, was chosen to be $ 0.081 $. The Polyak averaging coefficient was increased from $0.995$ to $0.999$, and the learning rate decayed from $5 \times 10^{-5}$ to $4 \times 10^{-6}$ over the course of training. In addition to the Q-value loss, we incorporated a minor L2 weight regularization term with a coefficient of $10^{-6}$, although this loss term may not be strictly necessary.

The latest network parameters were generated by the Learner every 15 seconds, and the network parameters of self-play Actors were updated every 5 minutes. In the self-play games of Actors, we enforced a maximum episode length of $19 \times 19 \times 2 = 722$ steps. If the step count reached this limit before the game concluded, the game was forcibly terminated, and a final reward was calculated. We added a small minimum probability for each action choice as an exploration noise. The size of Replay Buffer was initially set to $3 \times 10^8$; however, considering the trade-off between buffer size and data staleness, we ultimately reduced it to $1.5 \times 10^8$. Neural network training was performed using multi-GPU data parallelism with a batch size of 256 per GPU, resulting in a total training batch size of $256 \times 6 = 1,536$.

All the hyperparameters used in the training process are summarized in Table~\ref{tab:hparams}.

\begin{table}[h]
\centering
\begin{tabular}{c|c}
\hline
\;\;\;\;\;Hyperparameters/detail\;\;\;\;\;
&  \;\;\;\;\;\;\;\;\;\;\;\;\;\;value\;\;\;\;\;\;\;\;\;\;\;\;\;\;  \\\hline
Temperature $\alpha$             &  0.081  \\ 
L2 weight regularization $c$      &  0.000001  \\ 
Polyak averaging $\rho$          &  0.999 \\ 
Discount factor $\gamma$          &  1.0 \\ 
Learning rate                    &   $\{ 10^{-5}, 10^{-6} \}$  \\ 
Max episode length        &  722 \\ 
Min action probability       & $ 3 \times 10^{-5}$ \\ 
Replay buffer size        & $ \{ 3.0, 1.5 \}  \times 10^8$ \\ 
Batch size                &  $1,536 $\\ 
Selfplay hardware         & 1 $\times$ 2080Ti GPU\;\, \\ 
Training hardware         & 6 $\times$ 2080Ti GPUs \\\hline
\end{tabular}
\caption{\label{tab:hparams}Hyperparameters and training details }
\end{table}

\subsection{Results}

Here we present an overview of the evaluation results of QZero. For simplicity, all evaluations in our experiments are based on the raw neural network, without using test-time compute such as test-time MCTS in AlphaGo \cite{AlphaGo}.

\begin{figure}[h]
\centering
\includegraphics[width=0.95\linewidth]{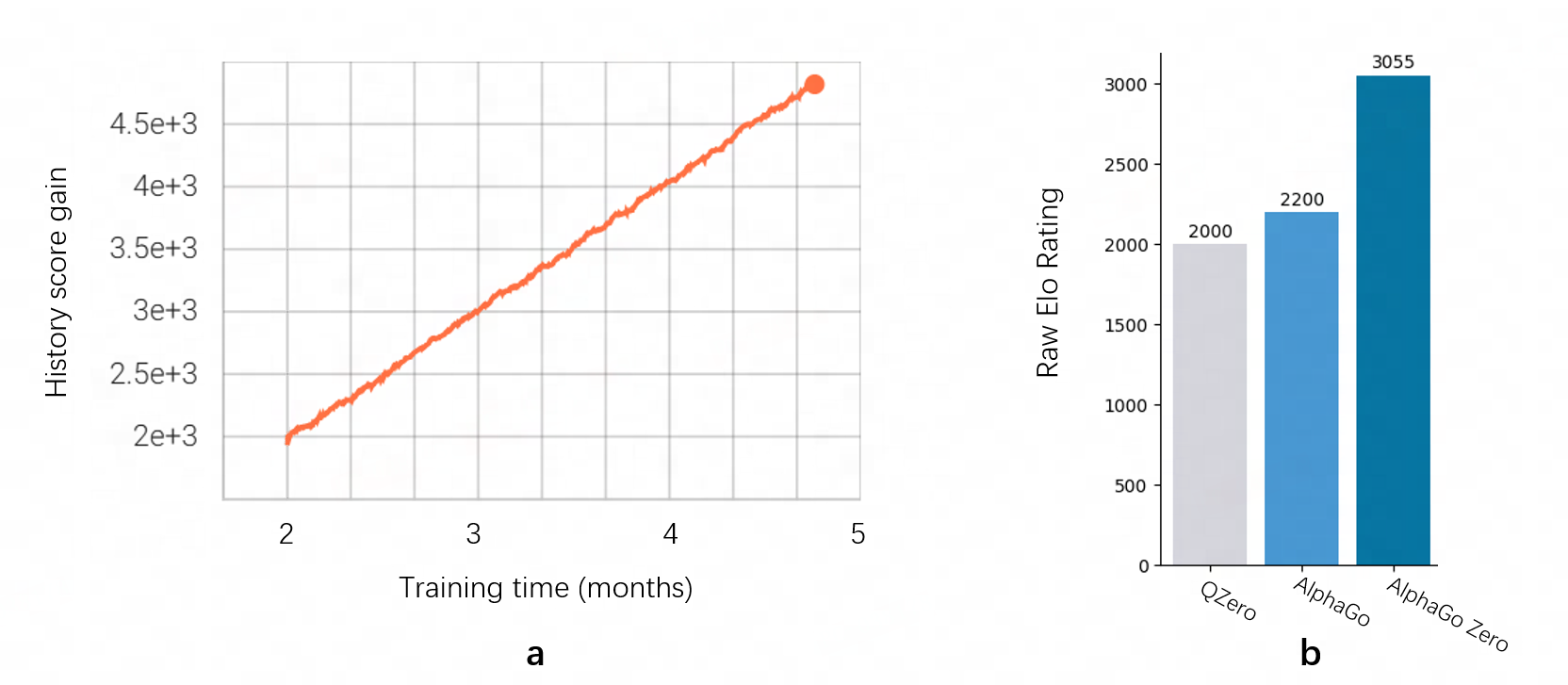}
\caption{\label{fig:curve}Evaluations.  \textbf{a.} The history score gain curve of QZero training.  \textbf{b.} Elo ratings of raw neural networks of QZero, AlphaGo and AlphaGo Zero.}
\end{figure}

During training, we utilize evaluators to track the agent's learning progress (see Figure~\ref{fig:arch}). The evaluators continuously update the win rates of network parameters in the history pool against the latest network parameters. For each historical network, we maintain a queue of length 10 to record the win/loss outcomes of matches against the latest network (1 if the latest network win else 0). Calculating the mean of the data in this queue yields the History Score for that specific historical network. We sum the history scores of all current historical network over time to obtain the curve of History Score Gain (HSG). This evaluation method is simpler and less computationally expensive than calculating Elo ratings \cite{elo}. 
As shown in Figure~\ref{fig:curve}\textbf{a},
the HSG curve is highly stable (nearly a straight line) and shows no signs of convergence or saturation, which indicates that the agent is still in a process of steady improvement. 
We originally planned to employ the $\alpha$-annealing schedule during training, specifically by gradually decaying the temperature $\alpha$ as the learned policy approached a Nash Equilibrium. Since training was far from convergence, we did not execute $\alpha$-annealing but kept $\alpha$ unchanged. We intend to experiment with it based on future training status.

After about 5 months of training, we tested QZero's performance on an online player-versus-player Go platform. The platform has a rank system (from low to high: 18K, 17K, ..., 1K, 1D, 2D, ..., 9D), where players are matched according to their ranks. 
During the online evaluation, we select the move with the maximum Q-value output from the raw neural network. Games were played with human players with 7.5 komi, no handicap, and Chinese rules. Finally, QZero's rank settled at 5D, and the Elo rating for 5D on this platform is estimated to be $2,000\sim2,100$ (see Figure~\ref{fig:curve}\textbf{b}).
In comparison, the Elo rating of AlphaGo without using MCTS as test-time compute is about 2,200 \cite{AlphaGo}, and the raw network of AlphaGo Zero achieves an Elo rating of 3,055 \cite{alpahgozero}. AlphaGo Zero was trained for 40 days using a 40 residual block neural network. Utilizing test-time compute can significantly enhance the strength of the raw neural network during evaluation. When combined with test-time MCTS, AlphaGo Zero achieves an Elo rating of 5,185. The Elo ratings of top human professional players are typically around $3,600\sim3,800$ \cite{gr}.

\begin{figure}[h]
\centering
\includegraphics[width=1.0\linewidth]{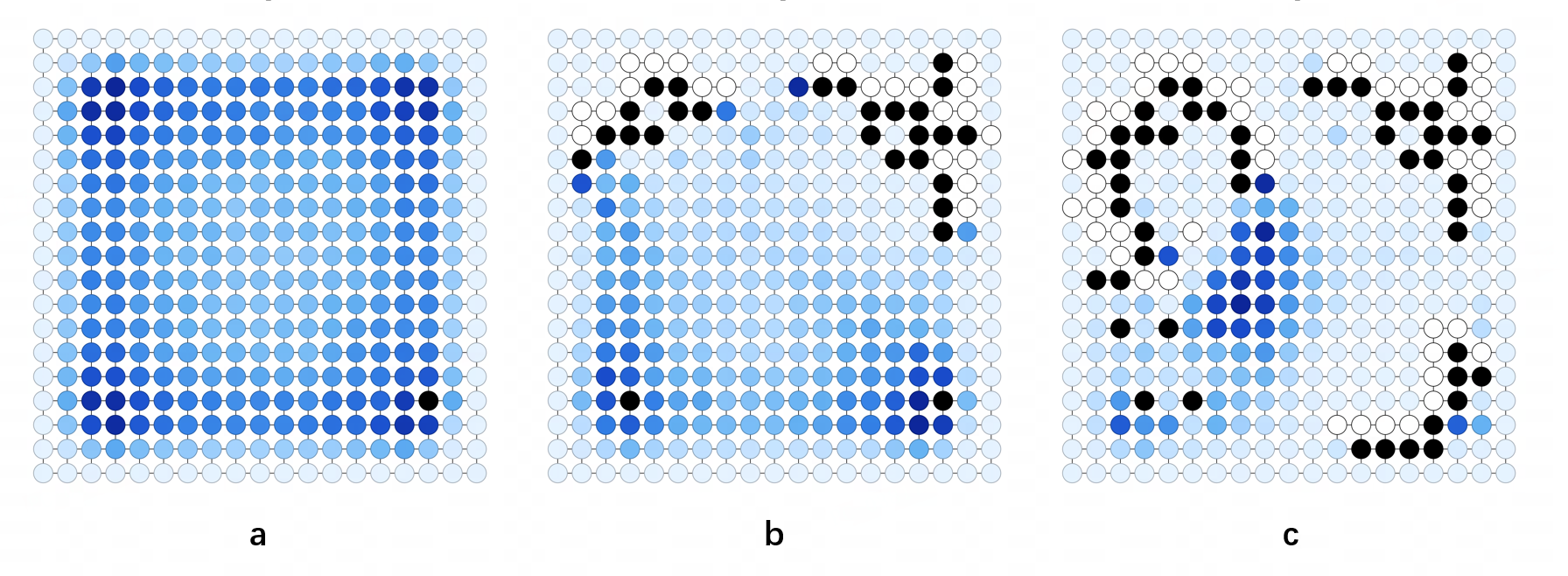}
\caption{\label{fig:prob}Action probability maps by the raw neural network of QZero. \textbf{a.} The probability map for the first move perfectly preserves the symmetries of the Go board. \textbf{b.} In the early stage of the game, the corners have higher value. \textbf{c.} As the game progresses, the AI gradually advances into the central region.}
\end{figure}

In this work, QZero's only goal during training is to learn a Nash equilibrium policy of the game of Go. 
Starting from completely random moves, QZero gradually developed a sophisticated understanding of the game of Go. Key concepts like life-and-death, ko, territory and influence, tactics in opening, middle game and endgame, are all learned from purely self-play reinforcement learning. 
In Figure~\ref{fig:prob}, we illustrate how moves are chosen in self-play games (from White's perspective). 
First, Q-values for each action are obtained through the inference the Q-value network. Subsequently, a probability distribution is generated based on the policy formula outlined in Algorithm~\ref{alg:QZero}. We utilize color intensity to visualize probability magnitude, where darker shades indicate higher probabilities.

\begin{figure}[h]
\centering
\includegraphics[width=1.0\linewidth]{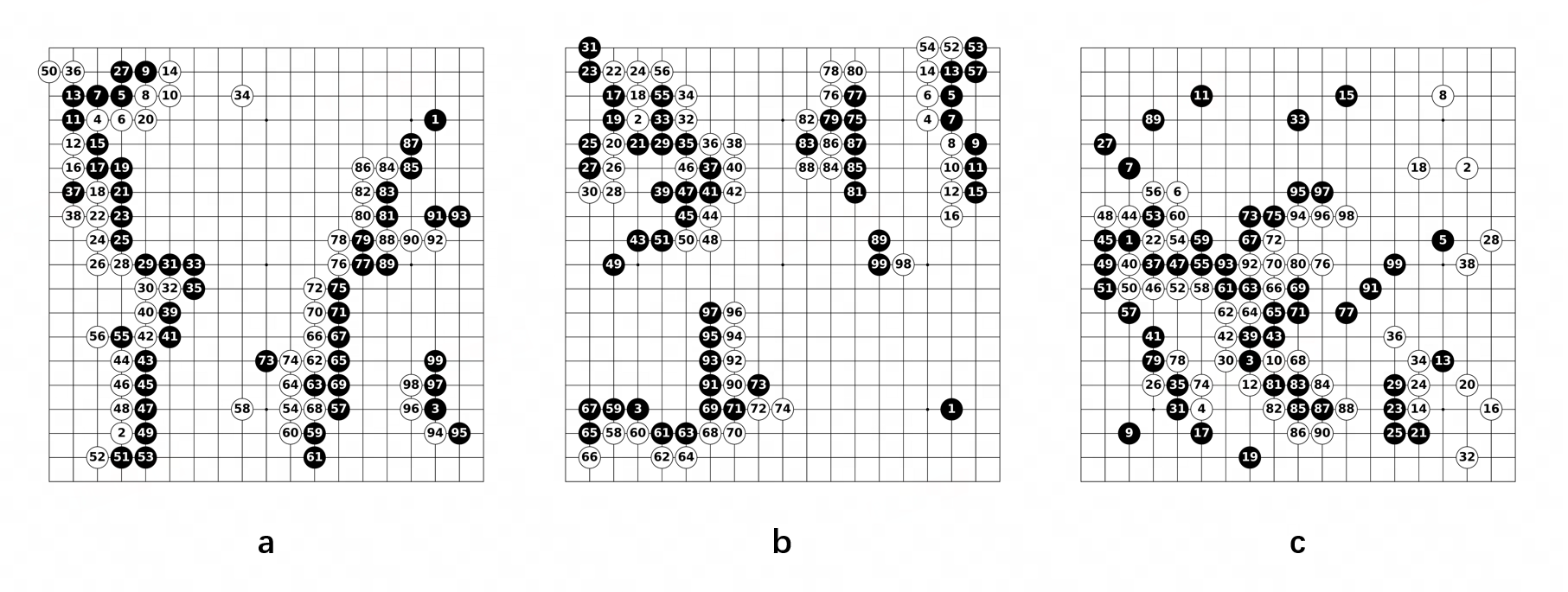}
\caption{\label{fig:games}Self-play games of the raw neural network of QZero. \textbf{a.} Game generated using argmax with the neural network parameters from the 3rd month. \textbf{b.} Game generated using argmax with the neural network parameters from the 5th month. \textbf{c.} Game generated using probabilistic sampling.}
\end{figure}

Given the probability distribution of the policy, we can select actions using either argmax or through probabilistic sampling. During training, sampling is exclusively used, including for the Actors' self-play and games played by the Evaluators. 
We utilize argmax for action selection only during online matches against human players or when testing the AI's performance.
In Figure~\ref{fig:games}, we showcase several self-play games by QZero.
The first two games use argmax, while the third employs sampling, resulting in a game that appears highly random. The last two games are generated with the same neural network parameters. Notably, the training data consists of games like the third game---such ``random data''---rather than the ``good data'' shown in the first two games. 
It is remarkable that QZero can learn such sophisticated skills from such stochastic data.

\section{Discussion}

In this section, we discuss several issues that are either unresolved in our experiments or beyond the scope of our experiments. 
It is hoped that this discussion will be instructive or inspiring for future research.

The temperature $\alpha$ regulates the balance between exploration and exploitation during training.
The exploration-exploitation trade-off is a fundamental issue in reinforcement learning. We can approach it by drawing an analogy from human learning.
Evidence from cognitive psychology suggests that an appropriate ratio of investment in new versus old knowledge is essential for achieving maximum learning efficiency. 
Thus, we argue that there exists a moderate $\alpha$
that maximizes learning efficiency. An excessively small 
$\alpha$ may result in squandering resources on the repetitive consolidation of existing knowledge, while an excessively large 
$\alpha$ may cause the learning signal to be overwhelmed by noise. 
Although we did not tune $\alpha$ in our experiments, we believe it is worthwhile to determine its optimal value experimentally before formal training.

QZero employs entropy regularization to stabilize the training process, thereby enhancing training efficiency. In an entropy-regularized self-play reinforcement learning environment, the Entropy Cancellation Approximation appears to be valid. Under this approximation, Eq.(\ref{eq:a7}) can be simplified to backup solely with Q-values.
We hypothesize that training without entropy regularization is feasible, though training stability and efficiency will be compromised.

Model-free and model-based RL are the two branches of RL divided by whether an environment model is used. Their policy improvement relies on learning and search (or planning), respectively.  
In this work, we demonstrate that learning-based methods, much like search-based methods, are capable of mastering games as complex as Go. A natural and interesting question arises: 
which of the two---learning or search---is more computationally efficient?
First, it is indisputable that learning methods are far superior to search methods in terms of data efficiency. Regarding computational efficiency, however, it is less obvious which method is better. 
QZero shifts the computational budget originally allocated for search in AlphaGo to learning. 
It is possible that one of them is superior, or that their efficiencies are comparable if it is the case that there is no free lunch and the strength of an agent is ultimately determined by the amount of compute used in training. We encourage interested readers to seek the answer to this question if their resources permit.

\section{Conclusion}

In this work, we presented QZero, a simple yet efficient model-free off-policy reinforcement learning algorithm. Theoretically, it can be viewed as a self-play version of the SQN algorithm \cite{sqn}. 
Starting tabula rasa without human data or model-based search, QZero learns a Nash equilibrium policy solely through self-play experience replay. Despite utilizing only 7 GPUs---a fraction of the computational resources employed by AlphaGo and its successors---our agent achieved a performance level comparable to AlphaGo.
Our findings highlight the efficiency of large-scale off-policy reinforcement learning methods.

Finally, 
the paradigm of intelligent agents learning through experience replay closely resembles human learning: the human brain learns by leveraging its own experiences, including those self-generated (such as in dreams). To tackle increasingly complex environments and develop more intelligent agents, the scalability and efficiency of reinforcement learning algorithms have been a relentless pursuit \cite{ssr}. In the Era of Experience \cite{ss}, it is hoped that our work will inspire future research in the fields of large-scale off-policy reinforcement learning and continual learning.
% and we hope they will inspire future research on more scalable, efficient, and general algorithms in reinforcement learning.

\bibliographystyle{alpha}
% \bibliography{sample}

% \bibliographystyle{apalike}
\bibliography{sample.bib}

\end{document}